\documentclass[12pt]{article}

\usepackage[utf8]{inputenc}
\usepackage{amssymb, amsmath}
\usepackage{graphicx}
\usepackage{subcaption}
\usepackage{float} 
\usepackage{xcolor}
\usepackage{booktabs} 
\usepackage{mathrsfs}
\usepackage{geometry}
\usepackage{natbib}
\usepackage{hyperref}

\newtheorem{definition}{Definition}

\title{\textbf{Fractional Artificial Neural Networks for Growth Models}}

\author{
	\begin{tabular}{c}
		Juan Carlos Nájera-Tinoco$^{1}$, Martín P. Arciga-Alejandre$^{1}$,\\
		Jorge Sánchez-Ortiz$^{1}$, Francisco J. Ariza-Hernández$^{1}$\thanks{Corresponding author: \texttt{arizahfj@uagro.mx}} \\[1.2em]
		\small $^{1}$\textit{Facultad de Matemáticas, Universidad Autónoma de Guerrero,}\\
		\small \textit{Av. Lázaro Cárdenas S/N, Chilpancingo de los Bravo, Guerrero 39087, México}
	\end{tabular}
}

\date{}

\begin{document}
	
	\maketitle
	
	\begin{abstract}
		In this paper we present a method to solve initial value problems for fractional growth models, such as generalizations of the exponential and logistic with periodic harvesting models. Using a discretization of the Caputo derivative we propose a fractional artificial neural network, which is implemented in the statistical software R. Moreover, we show examples where the analytical solutions and the approximation of the artificial neural network are compared.
	\end{abstract}
	
	
	\section{Introduction}
	
	Fractional derivatives are integro-differential operators that generalize the ordinary derivative \citep{Samko}, which have gained great popularity since fractional differential equations assume an important role in modeling the anomalous dynamics of many processes related to systems in science and engineering \citep{Podlubny, Trujillo}. This is due to the fact that the fractional derivative exhibits the property of nonlocality; i.e., the dynamics of the processes have long memory \citep{Luchko}.
	
	In the study of solutions for initial value problems involving fractional differential equations, numerical methods are a very important tool, since analytical solutions often cannot be found \citep{Garrappa}.
	
	The purpose of this paper is to present a technique to solve fractional differential equations by using artificial neural networks, which have been used in different fields of science and engineering due to their robustness and ability to provide practical solutions for real complex problems \citep{Pakdaman}. Also, the structure of the neural network leads to accurately describe complex models, where with simple efforts, the consumption of training time is reduced \citep{Romero}. On the other hand, \citet{Cybenko} stated that any continuous function of $n$ real values can be approximated by combinations of sigmoidal or smooth functions; as a consequence, neural networks have been used to solve initial value problems for ordinary differential equations \citep{Li}. 
	
	Artificial neural networks have also been used to solve fractional differential equations; for example, \citet{Qu} presented a method to solve fractional differential equations with initial value by algorithms using gradients. Also, \citet{Jafarian} used unsupervised back propagation learning algorithms to adjust the weights of neural networks and an appropriate truncated power series of the solution function to solve a class of fractional order initial value problems over a bounded domain. 
	
	In this work, we consider classical growth models, where we find their solutions using neural networks, which are programmed in R (statistical software) using gradient techniques. Moreover, we study a generalization of the growth models, where using a similar idea we find solutions using fractional artificial neural networks; that is, we use a discretization of the Caputo derivative to implement the artificial neural network. This technique is used to solve fractional growth models, where approximations of the solution for different values of the derivative order are obtained.
	
	\section{Solution Methods}
	
	\begin{definition}
		Let $x(t)$ be an absolutely continuous function in $[a,b]$ and $0<\alpha\leq 1$. Then the Caputo fractional derivative of order $\alpha$ is defined as:
		\begin{equation}
			D^{\alpha}x(t) = \frac{1}{\Gamma(1-\alpha)}\int_0^t \frac{x'(\tau)}{(t-\tau)^{\alpha}} \, d\tau,
		\end{equation}
		where $\Gamma(\cdot)$ is the Gamma function.
	\end{definition}

Now, if we have a Fractional Differential Equation (FDE) with initial conditions given by 
\begin{equation}\label{EDF}
	\begin{aligned}
		D^{\alpha}u(t) &= f(u,t), \ \ \ t \geq 0, \\
		u(0)&=u_0,
	\end{aligned}
\end{equation}

since Fractional Artificial Neural Networks (FANN) are good approximators, let's use this to approximate the solution of the FDE, so that
\begin{equation*}
	FANN(t) \approx u(t),
\end{equation*}

with this, we can say that the derivative of the fractional neural network will give us a similar equation
\begin{equation*}
	D^{\alpha}(FANN(t)) = f(FANN(t),t),
\end{equation*}

if we have a true approximation, then we could say that the derivative will also be a good approximation to the derivative of the true solution, such that
\begin{equation*}
	D^{\alpha}(FANN(t)) \approx f(u,t), \ \ \  t \geq 0.
\end{equation*}

With this, we can introduce our loss function, as we have the given derivative function $f(u,t)$ and we can calculate the derivative of the fractional neural network $D^{\alpha}FANN(t)$ at each step. With this, the following loss function is proposed using the root mean square error of the values, given by 
\begin{equation*}
	L=\sqrt{\sum_i \left[ D^{\alpha}FANN(t_i)-f(u(t_i),t_i)\right]^2}.
\end{equation*}

Then, we are going to include the initial condition in the loss function as follows, we define a new function given by
\begin{equation*}
	g(t)=u_0 + tFANN(t),
\end{equation*}

so, we can train  g(t)  to satisfy the FDE  instead of the Fractional Neural Network. Then, it will automatically be a solution to the derivative function. We can incorporate this new idea into our loss function:
\begin{equation}
	L=\sqrt{\sum_i \left[D^{\alpha}g(t_i)-f(u(t_i),t_i)\right]^2}.
\end{equation}

On the other hand, we need to discretize the Caputo derivative in order to implement it in our FANN, let's consider the following

\begin{equation}\label{Caputo}
	D^{\alpha}x(t_n) = \frac{1}{\Gamma(1-\alpha)}\int_{t_0}^{t_n} \frac{x'(\tau)}{(t_n-\tau)^{\alpha}} \ d\tau,
\end{equation}

now, let's discretize Caputo's derivative. Using backward difference for derivative of the function $x(t)$, we obtain
\begin{equation}\label{2h}
	x'(t)=\dfrac{x(t)-x(t-h)}{h}+\mathcal{O}(h^{1}),
\end{equation}

where $\mathcal{O}$ is the Landau symbol. Now, substituting (\ref{2h}) in (\ref{Caputo})  and using the partition $0=t_0<t_2<t_4< \dotsb <t_{k}=t $, for $k=0,1, \dotsb , n$ with a $kh=t_{k}$, where $h$ is the step size, we solve the following integral

\begin{align*}
	&\frac{1}{\Gamma(1-\alpha)}\int_{t_0}^{t_n} \frac{x'(\tau)}{(t_n-\tau)^{\alpha}} \ d\tau\\ &=\frac{1}{\Gamma(1-\alpha)}\sum_{k=0}^{n-1}\int_{t_{k}}^{t_{k+1}} \left( \frac{x(t_{k+1})-x(t_{k})}{h} + \mathcal{O}(h^{1}) \right) \dfrac{1}{(t_n-\tau)^{\alpha}} \ d\tau   \\
	&= \frac{h^{1-\alpha}}{h\Gamma(2-\alpha)} \sum_{k=0}^{n-1}\left( x(t_{k+1})-x(t_{k})\right)    
	\left[ (n-k)^{1-\alpha}-(n-k-1)^{1-\alpha} \right] + \mathcal{O}(h^{2-\alpha})\\
	&=\frac{1}{h^{\alpha}\Gamma(2-\alpha)}\sum_{k=0}^{n-1} [x(t_{k+1})-x(t_{k})]\Delta_{n,k} + \mathcal{O}(h^{2-\alpha}),
\end{align*}
where $\Delta_{n,k}=(n-k)^{1-\alpha}-(n-k-1)^{1-\alpha}$. Finally, we get 

\begin{equation}
	D^{\alpha}x(t_n) = \frac{1}{h^{\alpha}\Gamma(2-\alpha)}\sum_{k=0}^{n-1} [x(t_{k+1})-x(t_{k})]\Delta_{n,k} + \mathcal{O}(h^{2-\alpha}).
\end{equation}

The purpose of the above formula is to be implemented in our code, in this way we will obtain a fractional neural network, details of this will be provided later.\\

Neural networks are a type of artificial intelligence inspired by the structure and functioning of the human brain. They are composed of basic units called neurons; we can consider each neuron as a node. Artificial neural networks are formed by a set of neurons that we group in layers; the input layer receives initial data and the output layer produces the final result of the process. All intermediate layers other than the input and output layers are called hidden layers, these layers process the information received.\\

Each neuron receives a series of inputs that we will note as $x_i$, which are multiplied by a weight $w_{ij}$, this being the weight associated with the synopsis connecting the $i-th$ input to the $j-th$ neuron. Occasionally, we will be interested in the behavior of the neuron being biased. The bias exerts, in an artificial neuron, the role of the loss currents of a biological neuron \citep{LeCun}. Mathematically, we can add this bias as an internal parameter that depends on the state of the neuron, i.e.
\begin{equation*}
	y_{ij}=\sum_{i=1}^{n} x_i w_{ij} + b_j,
\end{equation*}
where $y_{ij}$ represents the output of each neuron.\\

Although so far we have assumed that the combination of inputs and weights directly generated the output, in the general model of an artificial neuron, the generation of the output of the neuron is the responsibility of an activation phase independent of the input integration phase. The activation phase consists of applying some kind of nonlinear transformation to our net input $y_{ij}$, i.e.
\begin{equation*}
	z_j= \phi(y_{ij}),
\end{equation*}

where $\phi(\cdot )$ is the activation function of the neuron, usually nonlinear, there are some candidate activation functions $\phi(\cdot )$ such as rectified linear unit, sigmoid function, hyperbolic tangent, among many others \citep{Lederer}.\\

Once the value $z_j$ is obtained, it is used to classify the input $x_i$. If $z_j$ is above the neuron activation threshold, the neuron is activated and decides that the input vector corresponds to an input to a target class, usually a positive class (numeric value =1), otherwise the input vector does not cause the neuron to be activated and it is associated with the negative class (numeric value = 0). This is where an adjustment of the weights in the neural network is performed, using a training algorithm. The training of a multilayer network is usually performed using a propagation algorithm. This allows the error gradient to be calculated with respect to the different parameters of the network. Combined with an optimization technique such as Adam (Adaptive Moment Estimation), which combines the advantages of other optimizers such as stochastic gradient descent (SGD) with momentum and adaptive learning optimization \citep{LeCun}, a training algorithm for multilayer networks is obtained.\\

The gradient is calculated for an error function, also known as a loss function, and the optimization method adjusts the weights of the network with the objective of minimizing that error or loss function.

\section{Models and results}

In this section we observe the performance of the neural network to approximate the solution of some classical growth differential equations, we evaluate the results of the network and the analytical solutions, then, we analyze the behavior of the method proposed in the previous section to solve fractional growth differential equations, and we show results for different values of alpha, in addition we will show the behavior of the loss function of each of the exposed results.\\

\textbf{Example 1: Linear growth model}\\

\begin{figure}[H]
	\centering
	\includegraphics[width=0.8\textwidth]{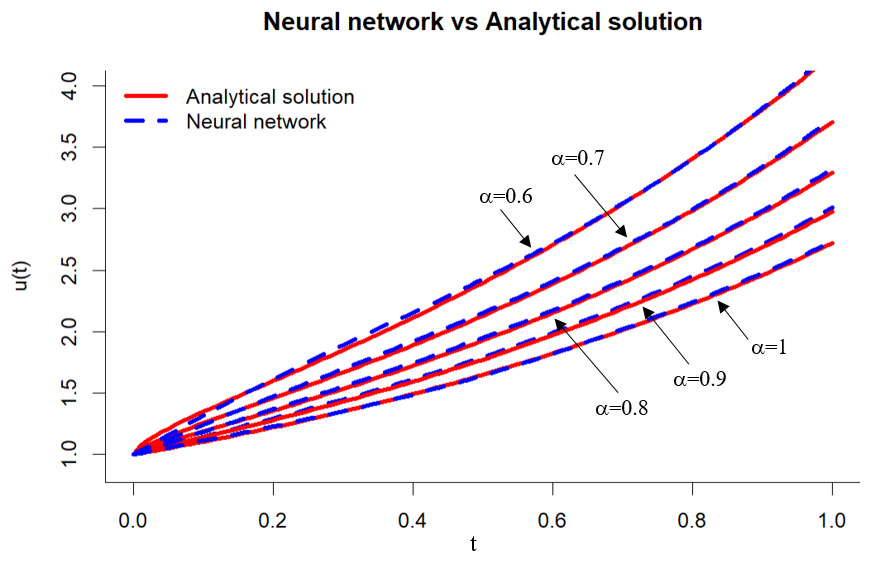}
	\caption{Fractional exponential growth.}
	\label{fig:Fractional exponential}
\end{figure}

For our first example, we take the ordinary and fractional exponential growth equation, which has a known analytical solution, with this we can compare that solution with the proposed fractional neural network approximation, we show the results for different values of alpha and then the loss function for each alpha is observed.
The first differential equation we take is the one described by 
\begin{equation}
	\begin{aligned}
		D_t^{\alpha}u(t) &= au(t), \\
		u(t_0) &= u_0,
	\end{aligned}
\end{equation}
where $a$ is the growth rate, if $a>0$ it is growth and $a<0$ is decrease. The exponential growth equation with initial condition, the explicit solution is given by $u_0 E_{\alpha}(at^{\alpha})$. Using the neural network with one input layer, one output layer and two hidden layers, each with 42 neurons, for this example the sigmoid activation function was used, we can observe the following results using a growth rate $a=1$ and an initial condition $u_0=1$ in addition we show results for different values of the derivative see Figure \ref{fig:Fractional exponential}.

\begin{figure}[H]
	\centering
	\begin{subfigure}{0.44\textwidth}
		\centering
		\includegraphics[width=\textwidth]{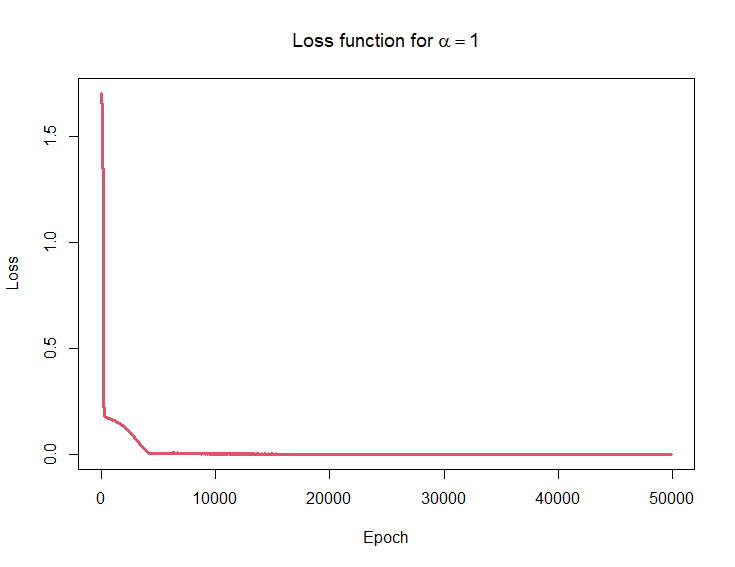}
	\end{subfigure}
	\hfill
	\begin{subfigure}{0.44\textwidth}
		\centering
		\includegraphics[width=\textwidth]{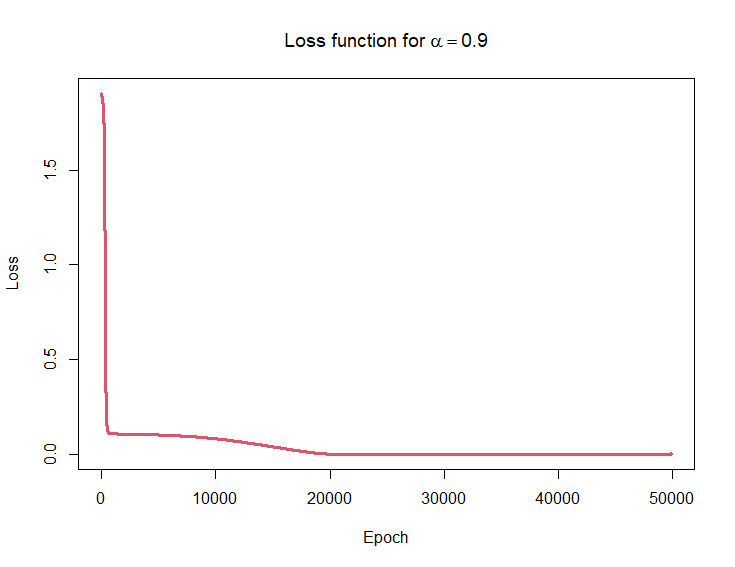}
	\end{subfigure}
	
	\vspace{0.2cm} 
	
	\begin{subfigure}{0.44\textwidth}
		\centering
		\includegraphics[width=\textwidth]{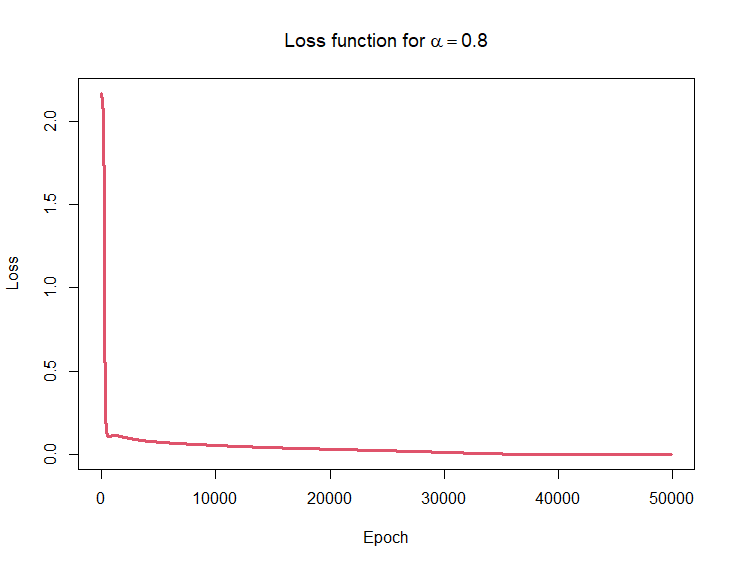}
	\end{subfigure}
	\hfill
	\begin{subfigure}{0.44\textwidth}
		\centering
		\includegraphics[width=\textwidth]{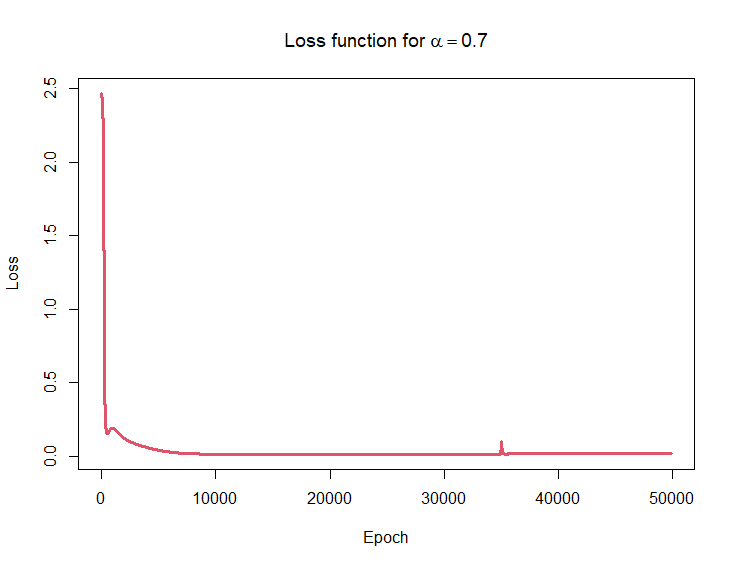}
	\end{subfigure}
	
	\caption{The loss functions for $\alpha= 1, 0.9, 0.8$ and $0.7 $ respectively. The $x$-axis represents the training epochs while the $y$-axis represents the loss.}
	\label{fig:Loss functions}
\end{figure}

\textbf{Example 2: Nonlinear growth model}\\

For our second example we use the fractional logistic growth equation, which is used in different scientific areas due to its ability to model physical phenomena with limited growth. If we take alpha = 1 the equation has a known analytical solution, but for alphas less than 1 there is no analytical solution, however with our proposed fractional artificial neural network we can find a numerical approximation to the solution. 

\begin{equation}
	\begin{aligned}
		D_t^{\alpha}u(t) &= au(t) \left(1- \frac{u(t)}{N} \right), \\
		u(t_0) &= u_0,
	\end{aligned}
\end{equation}
where $a$ is the growth rate, $N$ is the carrying capacity and $u_0$ is the initial condition, i.e., the value of $u$ in $t=0$.\\

\begin{figure}[htbp]
	\centering
	\includegraphics[width=0.9\textwidth]{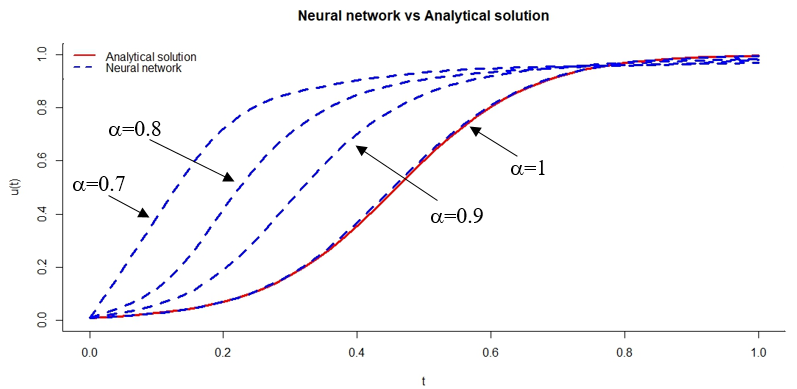}
	\caption{Fractional logistics growth.}
	\label{fig:Fractional logistics}
\end{figure}

By implementing an appropriate structure to the fractional neural network with one input layer, one output layer and six hidden layers, with 8, 42, 64, 64, 42, 8 neurons respectively, for this example the sigmoid activation function was used, the parameters used for the model are $N=1$, $u_0=0.01$ and a growth rate $a=10$, for different values of the derivative the following results were obtained see Figure \ref{fig:Fractional logistics}.  

\begin{figure}[H]
	\centering
	\begin{subfigure}{0.45\textwidth}
		\centering
		\includegraphics[width=\textwidth]{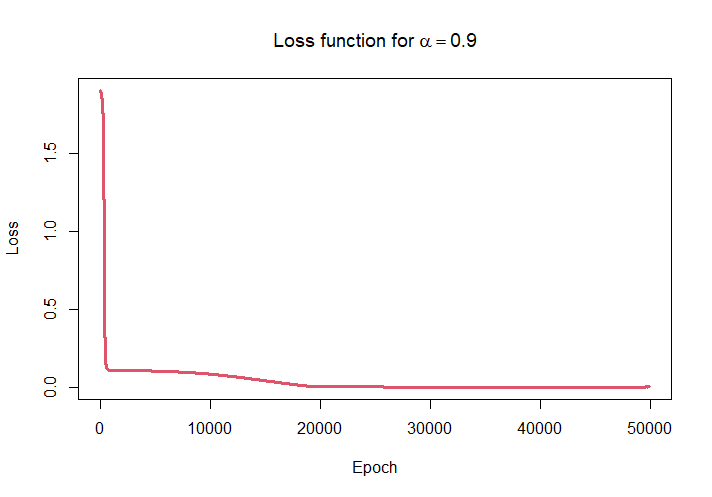}
	\end{subfigure}
	\hfill
	\begin{subfigure}{0.45\textwidth}
		\centering
		\includegraphics[width=\textwidth]{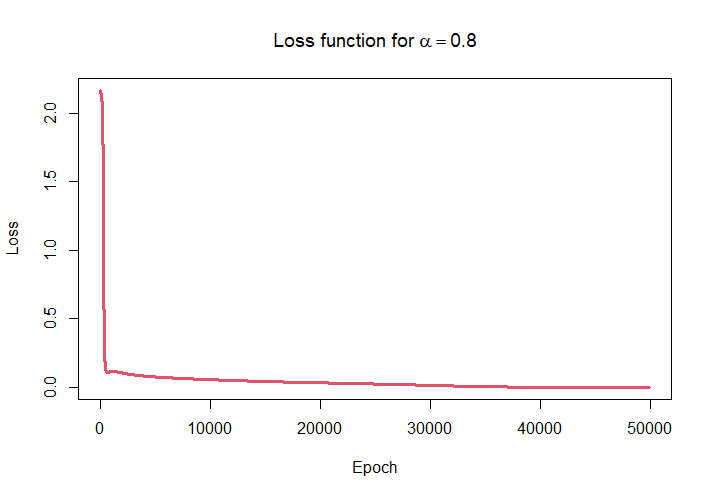}
	\end{subfigure}
	
	\vspace{0.2cm} 
	
	\begin{subfigure}{0.45\textwidth}
		\centering
		\includegraphics[width=\textwidth]{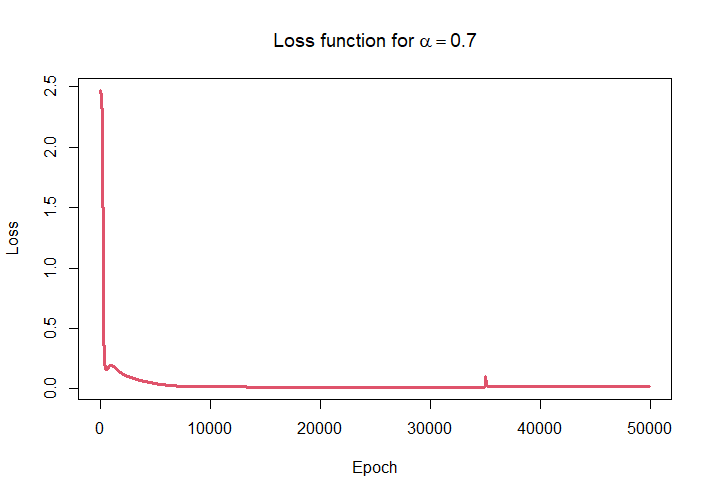}
	\end{subfigure}
	\hfill
	\begin{subfigure}{0.45\textwidth}
		\centering
		\includegraphics[width=\textwidth]{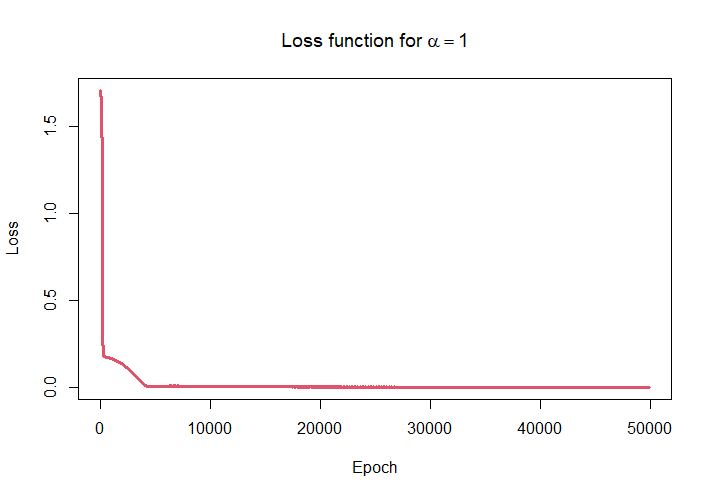}
	\end{subfigure}
	
	\caption{The loss functions for $\alpha= 1, 0.9, 0.8$ and $0.7 $ respectively.}
	\label{fig:Loss functions}
\end{figure}

\textbf{Example 3: Nonlinear growth model with periodic harvesting}\\

For our third example we will add a source to the logistic growth equation, 

\begin{equation}
	\begin{aligned}
		D_t^{\alpha}u(t) &= au(t) \left(1- \frac{u(t)}{N} \right) - b(1+\sin(2 \pi t)), \\
		u(t_0) &= u_0,
	\end{aligned}
\end{equation}
where $a$ is the growth rate, $N$ is the carrying capacity, $b$ is harvesting parameter  and $u_0$ is the initial condition, i.e., the value of $u$ in $t=0$.\\

\begin{figure}[h]
	\centering
	\includegraphics[width=0.9\textwidth]{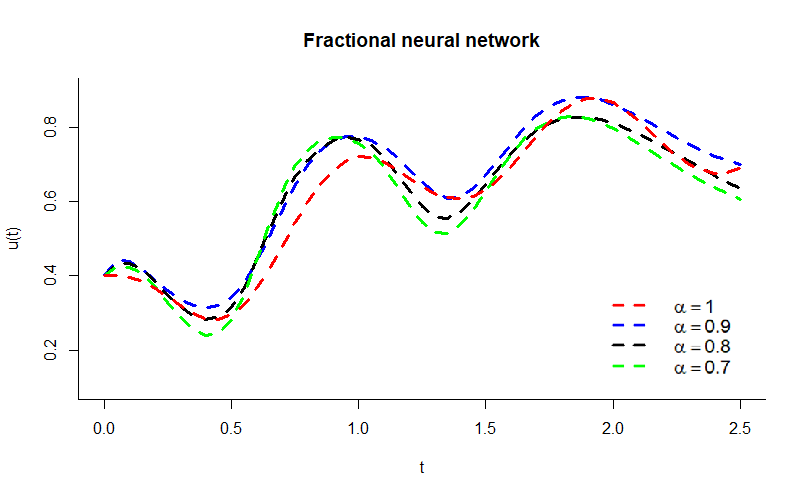}
	\caption{Fractional logistic model with periodic harvesting.}
	\label{fig:Fractional1}
\end{figure}

For this example in our neural network we use one input layer, one output layer and six hidden layers, with 8, 42, 64, 64, 42, 8 neurons respectively, for this example the sigmoid activation function was used,  the parameters used for the model are $N=1$, $u_0=0.4$, $b=0.8$ and a growth rate $a=5$, for different values of the derivative the following results were obtained see Figure \ref{fig:Fractional1}

\begin{figure}[H]
	\centering
	\begin{subfigure}{0.45\textwidth}
		\centering
		\includegraphics[width=\textwidth]{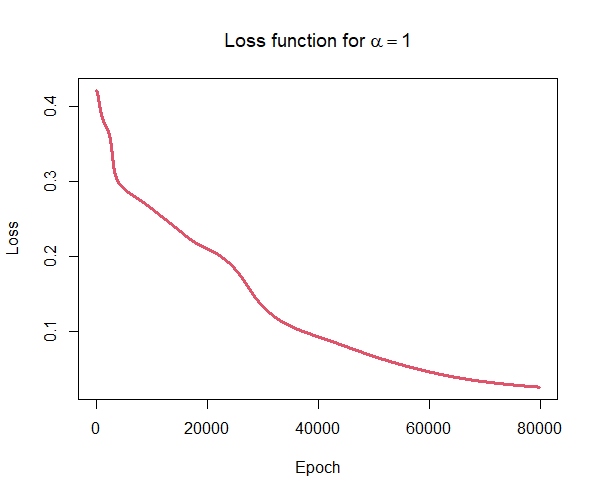}
	\end{subfigure}
	\hfill
	\begin{subfigure}{0.45\textwidth}
		\centering
		\includegraphics[width=\textwidth]{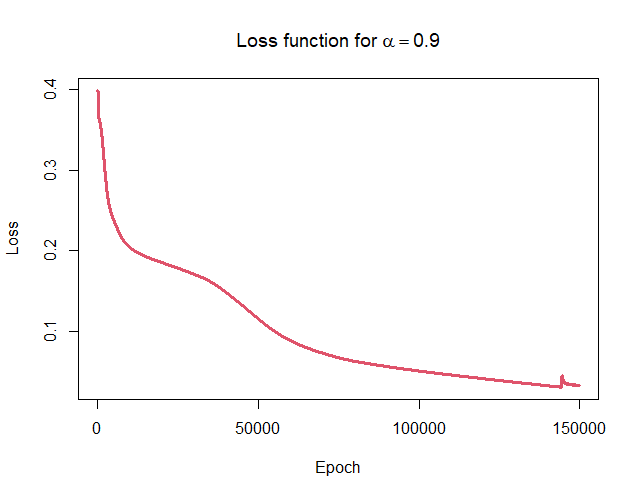}
	\end{subfigure}
	
	\vspace{0.2cm} 
	
	\begin{subfigure}{0.45\textwidth}
		\centering
		\includegraphics[width=\textwidth]{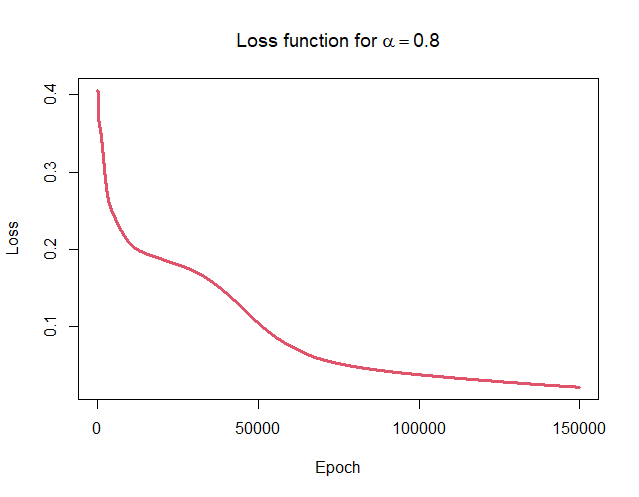}
	\end{subfigure}
	\hfill
	\begin{subfigure}{0.45\textwidth}
		\centering
		\includegraphics[width=\textwidth]{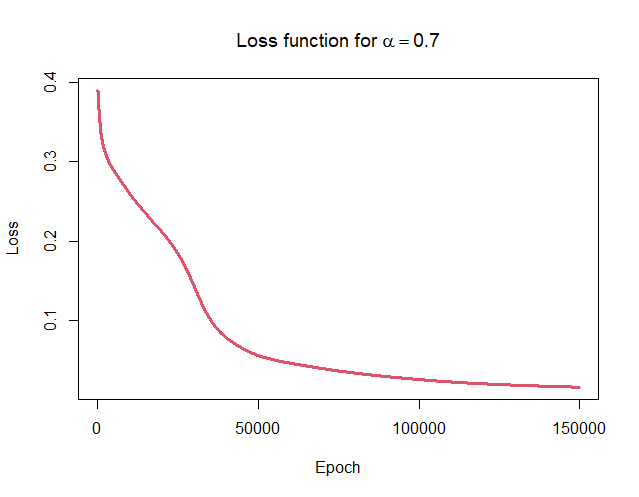}
	\end{subfigure}
	
	\caption{Loss functions for different alphas.}
	\label{fig:Loss functions}
\end{figure}
	
	\section{Conclusions}
	
	In this work, a fractional artificial neural network was presented to solve initial value problems in fractional growth models, using a discretization of the Caputo derivative. The results obtained in R show that the neural network achieves satisfactory approximations. The proposed methodology validates the use of neural networks for this type of problems and suggests its potential application in other fractional models. Finally, future research is aimed at improving the architecture of the neural network.

\end{document}